\documentclass[10pt,conference]{IEEEtran}
\IEEEoverridecommandlockouts

\usepackage{cite}
\usepackage{amsmath,amssymb,amsfonts}
\usepackage{algorithmic}
\usepackage{graphicx}
\usepackage{textcomp}
\usepackage{xcolor}
\def\BibTeX{{\rm B\kern-.05em{\sc i\kern-.025em b}\kern-.08em
    T\kern-.1667em\lower.7ex\hbox{E}\kern-.125emX}}

\usepackage{times}  
\usepackage{helvet}  
\usepackage{courier}  
\usepackage[hyphens]{url}  
\usepackage{graphicx} 
\usepackage{natbib}  
\setcitestyle{numbers,square}
\usepackage{caption} 
\usepackage{tabularx}
\usepackage{booktabs}
\usepackage{amssymb}
\usepackage{comment}
\usepackage{array}
\usepackage[table]{xcolor}
\usepackage{pifont} 
\usepackage{amsmath}
\usepackage{adjustbox}
\usepackage{multirow}
\usepackage{algorithm}
\usepackage{algorithmic}
\usepackage{xcolor}
\usepackage{newfloat}
\usepackage{listings}
\DeclareCaptionStyle{ruled}{labelfont=normalfont,labelsep=colon,strut=off} 
\floatstyle{ruled}
\newfloat{listing}{tb}{lst}{}
\floatname{listing}{Listing}
    
\begin{document}

\title{RAG-X: Systematic Diagnosis of Retrieval-Augmented Generation for Medical Question Answering}

\author{
\IEEEauthorblockN{Aswini Sivakumar,
Vijayan Sugumaran,
Yao Qiang}

Oakland University,
Rochester, MI, USA
\\
Email: \{aswinisivakumar, sugumara, qiang\}@oakland.edu
}

\maketitle

\begin{abstract}
Automated question-answering (QA) systems increasingly rely on retrieval-augmented generation (RAG) to ground large language models (LLMs) in authoritative medical knowledge, ensuring clinical accuracy and patient safety in Artificial Intelligence (AI) applications for healthcare. Despite progress in RAG evaluation, current benchmarks focus only on simple multiple-choice QA tasks and employ metrics that poorly capture the semantic precision required for complex QA tasks. These approaches fail to diagnose whether an error stems from faulty retrieval or flawed generation, limiting developers from performing targeted improvement. To address this gap, we propose RAG-X, a diagnostic framework that evaluates the retriever and generator independently across a triad of QA tasks: information extraction, short-answer generation, and multiple-choice question (MCQ) answering. RAG-X introduces Context Utilization Efficiency (CUE) metrics to disaggregate system success into interpretable quadrants, isolating verified grounding from deceptive accuracy. Our experiments reveal an ``Accuracy Fallacy", where a 14\% gap separates perceived system success from evidence-based grounding. By surfacing hidden failure modes, RAG-X offers the diagnostic transparency needed for safe and verifiable clinical RAG systems. 
\end{abstract}


\section{Introduction}

\noindent
The reasoning capabilities of Large Language Models (LLMs) have spurred their rapid integration into healthcare applications, such as clinical decision support \cite{gaber2025}, automated medical note generation \cite{wang-etal-2025-towards-adapting}, and medical education \cite{llm-med-edu2023}. However, the safe deployment of standalone LLMs is constrained by well-known risks such as hallucinations \cite{survey_hallucination}, outdated knowledge \cite{clusmann2023future}, and a lack of source verifiability \cite{wu2024well}. 

Recently, Retrieval Augmented Generation (RAG) has emerged as the standard architecture to mitigate these issues by grounding LLMs' responses in authoritative, up-to-date knowledge \cite{rag_4_nlp}. The RAG system effectively separates the task into two complementary components: a retriever, which searches external knowledge sources to gather relevant context, and a generator (the LLM), which leverages this context to produce an informed, factually accurate response. However, unlocking its full potential requires rigorous evaluation and systematic diagnosis of each component, ensuring that the system operates reliably and safely in critical medical domains.

\noindent
\begin{figure*}[htbp]
    \centering
    \includegraphics[width=0.98\linewidth]{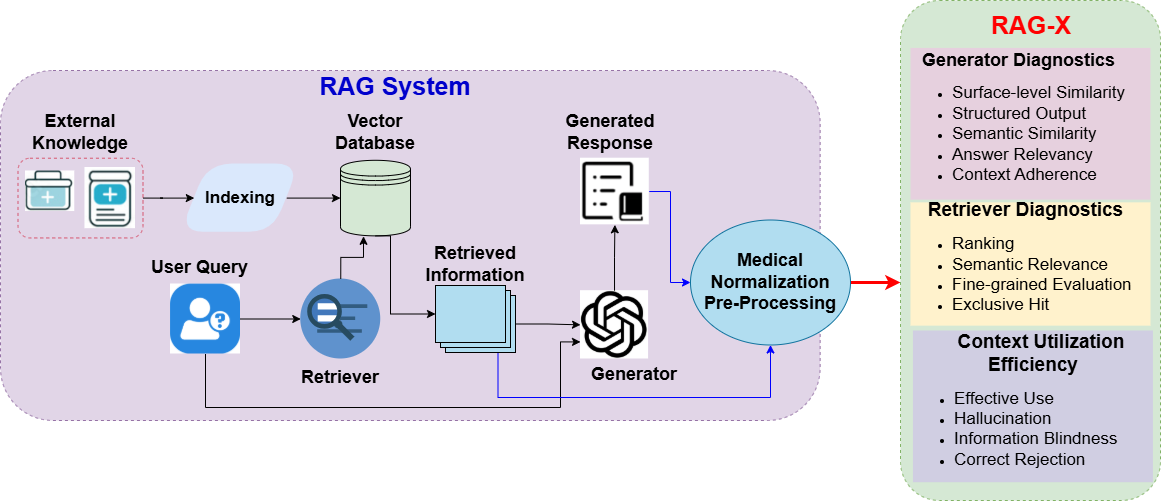}
    \caption{Overview of RAG System and RAG-X Diagnostics. This figure illustrates the workflow of an RAG system, where external knowledge is indexed into a vector database. A user query interacts with the retriever to fetch relevant information, which is then passed to a generator to produce a response. The RAG-X framework adds diagnostic modules for both retrieval (e.g., ranking, semantic relevance, fine-grained evaluation) and generation (e.g., surface-level similarity, structured output, semantic similarity, and LLM-based judgment) to provide detailed performance analysis.}
    \label{fig:overview}
    \vspace{-0.5 cm}
\end{figure*}

Although the evaluation of RAG systems is steadily advancing, existing frameworks suffer from a ``diagnostic" gap. Most prominent approaches rely on aggregate metrics such as accuracy or F1, which provide a high-level snapshot of system performance but fail to reveal the underlying causes of failure. As a result, developers cannot determine whether a wrong answer stems from a retriever that returned irrelevant context or a generator that misinterpreted relevant information. This limitation exposes a critical research gap in the medical domain: \textbf{the absence of evaluation frameworks that rigorously assess RAG systems on diverse, clinically realistic tasks while providing actionable, component-level diagnostics}.

To bridge this gap, we present RAG-X, a diagnostic framework that disaggregates system success into interpretable failure modes. Unlike conventional evaluations, RAG-X provides a multi-dimensional analysis of the retrieval-generation interface, as shown in Figure \ref{fig:overview}. For retrieval, RAG-X moves beyond standard ranking metrics, such as Mean Average Precision, to reveal structural inefficiencies. For instance, in our evaluation of a best-performing RAG pipeline, RAG-X identified that over one-fifth (22\%) of the retrieved evidence was redundant, effectively wasting the model's limited context window despite having an adequate recall rate.

The core innovation of RAG-X lies in its \textbf{Context Utilization Efficiency (CUE)} metrics, which synchronize retriever success with generator behavior to expose the \textit{Accuracy Fallacy}, a scenario where the system appears to be highly accurate but is actually ungrounded. CUE isolates critical attribution errors in which a model's high adherence score masks a lack of true source grounding. In our experiments, RAG-X uncovered a 14\% gap between overall accuracy and actual evidence-based grounding, showing more than one-third (33.9\%) of the responses were ungrounded ``lucky guesses". These insights are essential for determining whether a system is genuinely following clinical evidence or merely appearing correct without verifiable support.

We state our primary contributions as follows:
\begin{itemize}
    \item \textbf{RAG-X: A Unified Diagnostic Framework for RAG Systems.} We introduce a multi-dimensional evaluation framework that replaces aggregate metrics with component-level diagnostics tailored for the high-precision requirements of medical QA.
    \item \textbf{Context Utilization Efficiency (CUE).} Our framework systematically categorizes RAG outputs into diagnostic quadrants, such as Effective Use and Hallucination, isolating grounded successes from deceptive ``lucky guesses".   
    \item \textbf{Attribution Error Identification.} We provide a systematic method to identify the \textit{Adherence Paradox}, demonstrating how RAG-X exposes cases where high adherence scores mask a fundamental lack of source grounding.
    \item \textbf{Comprehensive Empirical Study.} We evaluate RAG-X across three clinical datasets, validating its ability to identify actionable bottlenecks that standard Accuracy and F1-scores obscure.
\end{itemize}


\vspace{-0.3 cm}
\section{Related Work}


\subsection{Application of RAG in Medical Domain}
\noindent
Recently, meta-analyses \cite{improving_LLM_meta_analysis} indicate that integrating retrieval mechanisms yields a 1.35 odds ratio over baseline LLM performance and can deliver accuracy improvements exceeding 18\% using advanced prompting alone. These gains have motivated the adoption of RAG systems in diverse healthcare applications, such as clinical decision support \cite{lu2024clinicalrag,gaber2025}, automated medical note generation \cite{wang-etal-2025-towards-adapting}, medical QA \cite{xiong-etal-2024-benchmarking}, and medical education \cite{llm-med-edu2023}. However, medical applications pose domain-specific challenges that differ significantly from those in general LLM applications. Medical information is inherently heterogeneous, encompassing unstructured clinical narratives \cite{luo2024zero}, semi-structured EHRs \cite{li2024scoping}, and highly structured guideline repositories \cite{levra2025large}. Clinical queries extend beyond basic fact retrieval and require "multi-hop/multi-step" reasoning \cite{krishna-etal-2025-fact}. This involves synthesizing information from multiple documents to reach a conclusion. Moreover, factual accuracy and source verifiability are non-negotiable due to the life-critical nature of clinical decisions \cite{enhancing_med_AI}. Errors arising from incomplete retrieval or hallucination can result in severe patient harm \cite{kim2025medical}, underscoring the need for specialized evaluation methods that assess robustness under these domain-specific constraints \cite{wu2024well}.

\subsection{Evaluation of RAG in Medical Domain}
\noindent
The need for robust evaluation has spurred the development of several RAG benchmarks. In the general domain, RAGAS \cite{es-etal-2024-ragas} has established a foundational reference-free approach to score answer faithfulness, context relevance, and retrieval precision using carefully designed heuristic prompts. However, these heuristic scoring prompts often fail to adapt when applied to new domains or corpora. To mitigate this, ARES \cite{saadfalcon2024ares} introduced a pipeline for fine-tuning lightweight LLM judges on synthetic data and applying Prediction-Powered Inference (PPI) to derive statistically reliable confidence intervals for its evaluation metrics. While this framework evaluates the RAG components separately, each cross-domain shift requires in-domain passages and few-shot query examples to reconfigure ARES's judges. This limits its direct applicability to specialized domains or information-extraction tasks, underscoring the need for domain-specific evaluation frameworks.   

In the medical domain, frameworks such as MIRAGE \cite{xiong-etal-2024-benchmarking} benchmark RAG systems against clinical QA datasets, focusing on the impact of different retrieval modules on target accuracy. MedRGB \cite{ngo2024medrgb} extended this to evaluate information sufficiency, multi-document integration, and robustness to factual errors. Although these frameworks mark significant progress, they still face two major limitations. First, most existing benchmarks, such as MIRAGE and MedRGB, primarily focus on multiple-choice or factoid-style QA. These rely on simple accuracy-based metrics and fail to capture the semantic precision and reasoning adequacy required for complex medical tasks such as guideline recommendation or evidence synthesis. Second, while general evaluation frameworks, such as ARES and RAGAS, provide component-level scores, they lack diagnostic granularity to identify why a retrieval failed. 

To address these evaluation issues, we propose RAG-X, the first diagnostic framework tailored to medical RAG systems. Unlike prior benchmarks, RAG-X decouples retrieval and generation performance, enabling fine-grained error attribution. Furthermore, it introduces multi-dimensional metrics that evaluate coverage, redundancy, faithfulness, and reasoning adequacy, offering actionable insights beyond a single score. Table~\ref{tab:Framework Comparison} compares RAG-X with existing frameworks, highlighting its unique ability to provide component-level diagnostics essential for safe and effective deployment of RAG in healthcare settings.

\begin{table}[t]
\scriptsize
\centering
\rowcolors{2}{gray!30}{white}
\begin{tabularx}{\columnwidth}{lXXXXX}
\toprule
\textbf{Capability} & \textbf{RAGAS} & \textbf{MIRAGE} & \textbf{MedRGB} & \textbf{RAG-X} \\ 
\midrule
Primary Focus & General  & Medical & Medical & Medical \\
Metric Engine & Heuristic Prompts  & Accuracy & Prompt Analysis & Multi-level \\
Domain Alignment & None  & None & None & Logic-aware \\
Diagnostic Decoupling & Limited  & Minimal & Limited & Full \\
Task Coverage & Broad  & MCQ  & MCQ  & Broad \\
\bottomrule
\end{tabularx}
\caption{Comparison of capabilities across RAG evaluation frameworks. RAG-X uniquely incorporates independent component analysis and medical logic normalization, capabilities absent from existing benchmarks.}
\label{tab:Framework Comparison}
\vspace{-0.5 cm}
\end{table}

\section{Method}



\subsection{RAG Pipeline and Medical Normalization}
\noindent
The RAG pipeline evaluated in this study follows the standard three-stage architecture (indexing, retrieval, and generation) augmented with a domain-specific normalization layer.

\noindent
\textbf{-- Indexing} 
transforms the structured text chunks from raw documents into high-dimensional embeddings using a pretrained embedding model. The embeddings are indexed using the Hierarchical Navigable Small World Indexing algorithm and stored in a vector database. This indexing method enables data to be stored in a multi-layer structure, making approximate nearest neighbor searches fast and efficient. 

\noindent
\textbf{-- Retrieval} 
searches for the most relevant segments from the vector database based on their proximity to the query vector. We employ a hybrid search strategy that integrates traditional lexical matching with semantic similarity search using the reciprocal rank fusion (RRF) algorithm. The contribution of the sparse and dense retrieval is controlled by a weighting parameter $\alpha$. This approach is particularly important in the medical domain, where effective retrieval must capture both rigid medical terminology match and the nuanced, conceptual nature of the clinical queries \cite{luo2019hybrid}.

\noindent
\textbf{-- Generation} 
integrates the user query with the retrieved evidence to produce a response that is both factually grounded and contextually aligned. This is achieved by constraining the model's reasoning to the provided evidence rather than relying solely on its internal knowledge.

\noindent
\textbf{-- Medical Normalization Pre-processing:} We applied a specialized normalization process to the retrieved contexts, generated responses, and ground truths to ensure robust evaluation. This includes medical abbreviations to full terms mapping (e.g., ``AAA" to ``abdominal aortic aneurysm"), standardizing age-related thresholds, and unifying gender-specific phrasing.

\subsection{RAG-X Approach}
\noindent
Traditional RAG evaluations often fall short in three key areas: (1) distinguishing component-level accountability (retriever versus generator errors), (2) incorporating semantic and human-aligned measures of relevance and factuality, and (3) providing transparency into retrieval failure modes such as coverage gaps and redundancy. RAG-X addresses these limitations through a comprehensive diagnostic framework that integrates quantitative information retrieval metrics, embedding-based semantic similarity, and LLM-as-a-judge evaluations. 
Specifically, RAG-X organizes metrics into three primary categories: retrieval diagnostics, generation quality, and context utilization efficiency to support fine-grained, component-level interpretability and systematic failure-mode analysis across the entire RAG pipeline.


\noindent
\textbf{-- Retrieval Diagnostic Metrics:}
Retrieval quality fundamentally constrains the overall performance of any RAG system, as the generator cannot produce accurate or comprehensive answers without access to relevant evidence. Consequently, robust evaluation of the retriever is essential for diagnosing pipeline bottlenecks and guiding system improvements. To achieve this, we categorize our retrieval diagnostic metrics into three complementary groups:

\noindent
(1) \textit{Ranking-based Metrics}, such as Recall@$k$, Mean Average Precision (MAP), Mean Reciprocal Rank (MRR), and Normalized Discounted Cumulative Gain (nDCG) 
to evaluate the ranking quality of the retrieved contexts and measure how effectively the retriever surfaces relevant evidence.  

\noindent
(2) \textit{LLM-based Context Relevancy} complements traditional ranking metrics by leveraging an LLM-as-a-judge to assign a context relevancy score (0.0 - 1.0) to each retrieved segment, assessing how well that context can potentially answer the question \cite{zheng2023judging}.

\noindent
(3) \textit{Fine-grained Retrieval Diagnostics} offer deeper insights into specific retrieval failure modes. RAG-X introduces a set of diagnostic metrics specifically designed to characterize the underlying retrieval dynamics and evidence distribution. Let $Q$ be the set of all queries in the evaluation set. For each query $q \in Q$, let $C_q = \{c_1, c_2, \ldots, c_k\}$ be the ranked set of $k$ retrieved contexts and $GT_q$ be the ground truth answer. We define a cascading binary relevance function $R(c_i, GT_q)$ as:
\begin{equation}
    R(c_i, GT_q) =
    \begin{cases}
    1, & \text{if } c_i \text{ contains correct answer for } q, \\
    0, & \text{otherwise.}
    \end{cases}
\end{equation}
To compute this relevance signal robustly, we implement a multi-level context-checking approach that checks for exact substring matches,  followed by token-level overlap ( \( \geq 0.80 \)), and finally sentence-level semantic similarity using embeddings ( \( \geq 0.75 \)). Using this function, we derive the following diagnostic metrics:
\begin{itemize}
    \item \underline{Context-$k$ Hit Rate} measures the percentage of queries where the ground truth appears in the $k$-th retrieved document, enabling analysis of retrieval performance across different ranks.
    \item \underline{No-Hit Rate} captures the percentage of queries for which none of the retrieved contexts contain the ground truth,
    providing a direct measure of retrieval coverage failures. A high rate signals a critical weakness in the retriever’s ability to locate relevant information within the corpus.
    \item \underline{Exclusive Hit Rate} reports the percentage of queries where the ground truth appears in only a single retrieved context, 
    indicating whether the retriever is providing unique sources of evidence or redundant variations of the same information.
\end{itemize}

\noindent
\textbf{-- Generation Quality Metrics} ultimately determines the usability and trustworthiness of a RAG system. Even when relevant evidence is retrieved, the generator must accurately synthesize this information to produce responses that are both correct and contextually grounded. Poor generation can undermine the benefits of high-quality retrieval, making it essential to evaluate this component thoroughly. To capture the multifaceted nature of generation quality, RAG-X organizes its generation metrics into three complementary dimensions:

\noindent
(1) \textit{Surface-Level Similarity} measures lexical overlap to assess how closely the generated answer matches the reference text at the word and phrase level. Common metrics in this category include Exact Match, Fuzzy Match, ROUGE-L, and Token-level F1 (see Appendix Table~\ref{tab:metric_definitions}). While useful for evaluating completeness, these metrics cannot capture paraphrasing or deeper semantic equivalence.

\noindent
(2) \textit{Semantic Similarity} detects whether the generated responses preserve the intended meaning of the reference text beyond superficial word overlap. To achieve this, we compute cosine similarity between sentence embeddings of the normalized answers, providing a semantic-level measure of alignment. 

\noindent
(3) \textit{Structured Output} evaluates the correctness of answers that require enumerating multiple elements, such as lists of risk factors or symptoms. Traditional token-level metrics often fall short in these scenarios, as they do not account for the structure and order of itemized responses. To address this, we introduce a List-Component F1-score, which splits the generated and reference responses into individual items and computes an F1-score based on their overlap, ensuring fair and accurate assessment of list-based outputs.

\noindent
(4) \textit{LLM-Judgment} leverages an LLM to evaluate two critical dimensions of the generated answer: answer relevancy, whether the response directly and completely addresses the user’s query, and context adherence, whether the response is factually grounded in the retrieved evidence \cite{cohen2025ragxplain}. This approach offers a human-aligned perspective on accuracy and factual consistency, serving as an essential complement to traditional numerical metrics.

\noindent
\textbf{-- Context Utilization Efficiency} framework synthesizes retriever and generator performance to assess how effectively the system utilizes retrieved evidence. By cross-referencing retrieval success determined through the relevance function with the generator's context adherence score (threshold \( \geq 0.7 \)), we categorize each question into four diagnostic quadrants: (1) Effective use occurs when the retriever successfully found the answer, and the generator correctly utilized it. (2) Information blindness describes cases where the retriever found the answer, but the generator failed to utilize the provided context. (3) Hallucination (Lucky Guess) refers to situations where the retriever failed to find the answer, but the generator produced a correct response by claiming adherence to irrelevant context or relying on internal weights. (4) Correct rejection captures instances where the retriever failed to find the answer, and the generator correctly exhibited low adherence to the irrelevant context.

\section{Experiment Settings}
\noindent
We demonstrate the diagnostic capabilities of RAG-X by: (1) evaluating various RAG configurations across diverse medical QA tasks; (2) comparing against non-RAG baselines; (3) performing ablations to validate pipeline configurations

\noindent
\textbf{Datasets and Knowledge Bases:} We evaluate RAG-X on three medical QA benchmarks, each representing a distinct modality and paired with a domain-specific retrieval corpus. \textit{PubMedQA} provides 1,000 labeled multiple-choice questions (yes/no/maybe) grounded in biomedical abstracts \cite{jin2019pubmedqa}. \textit{GuidelineQA} includes 59 expert-curated questions targeting information extraction from a corpus of 11 comprehensive United States Preventive Services Task Force (USPSTF) clinical guideline documents\cite{luo2024zero}. This dataset emphasizes high-precision retrieval and extraction. \textit{MedQuAD-GHR} contains 500 QA pairs drawn from MedQuAD \cite{benabacha2019question} evaluated against 1,084 genetic condition files scraped from MedlinePlus Genetics Home Reference (GHR).

\noindent
\textbf{Baselines:} 
To quantify the effectiveness of the retrieval step, we compare RAG against two non-retrieval baselines: (1) \textit{Direct Zero-shot} (prompts question only), testing the model's parametric knowledge; (2) \textit{Long-Context Zero-shot} (prompts question plus the entire document), evaluating the model's ability to perform in-context learning.

\noindent
\textbf{RAG Pipeline Configuration:} 

\noindent
-- Indexing: Documents were segmented into chunks of 1,024 tokens with an overlap of 100 tokens and stored in a Weaviate \cite{weaviate} vector database collection. We evaluated three embedding models: Qwen3-Embedding-8B \cite{qwen3embedding}, MedCPT-Article-Encoder \cite{jin2023medcpt}, and NV-EmbedQA-E5-v5 \cite{nvidia2024nvembedqa}.

\noindent
-- Retrieval: We employed a hybrid search strategy that integrated BM25 for lexical matching with vector-based similarity search using RRF. For all experiments, we retrieved the top $k = 3$ context chunks and varied the weighting parameter $\alpha$.  

\noindent
-- Generation: We utilized three instruction-tuned models for response generation: Llama-3.1-8B-Instruct \cite{dubey2024llama}, gemma-2-9b-it \cite{gemma_2024}, and Qwen2.5-7B-Instruct \cite{qwen2}. For RAG-X metrics, all-MiniLM-L6-v2 \cite{sentence-transformers-pretrained-models} sentence-transformers model was employed to encode semantic similarity, and DeepSeek-V3.1 \cite{deepseekai2024deepseekv3technicalreport} served as the LLM-judge for all qualitative evaluations.

\noindent
    


\begin{table}[t]
\centering
\caption{Performance comparison of backbone LLMs across different settings on the GuidelineQA dataset. We report \textbf{Token-level F1-Score} to measure lexical precision and recall, \textbf{Semantic Similarity} to capture correctness in meaning, and \textbf{Accuracy} for the fraction of correct answers. An answer is considered accurate if it matches the ground truth using at least one of the following criteria: (1) Exact Match, (2) Fuzzy Match, (3) List-Component F1 score $\geq 0.7$, or (4) Semantic Similarity score $\geq 0.7$}
\label{tab:guidelineQA_comparison}
\begin{adjustbox}{max width=\linewidth}
\begin{tabular}{@{}llccc@{}}
\toprule
\textbf{LLM} & \textbf{Setting} & \textbf{F1-Score} & \textbf{Semantic Similarity} & \textbf{Accuracy} \\
\midrule
                   & Direct Zero-Shot & 0.21 & 0.52 & 0.35 \\
Llama-3.1-8B-Instruct       & Long-Context     & \textbf{0.34} & 0.63 & 0.54 \\
                   & RAG              & 0.32 & \textbf{0.69 }& \textbf{0.71} \\
\midrule
                  & Direct Zero-Shot & 0.16 & 0.50 & 0.33 \\
gemma-2-9b-it     & Long-Context     & 0.51 & 0.70 & 0.61 \\
                   & RAG              & \textbf{0.56} & \textbf{0.75} & \textbf{0.69} \\
\midrule
                    & Direct Zero-Shot & 0.31 & 0.60 & 0.50 \\
Qwen2.5-7B-Instruct & Long-Context     & 0.27 & 0.52 & 0.47 \\
                   & RAG              & \textbf{0.44} & \textbf{0.68} & \textbf{0.54} \\
\bottomrule
\end{tabular}
\end{adjustbox}
\label{tab:llm_eval}
\end{table}


\section{Results and Discussions}
\vspace{-0.05em}
\noindent
\subsection{Comparison of Backbone LLMs}
\noindent
We first evaluated three backbone LLMs, i.e., Llama-3.1-8B-Instruct, gemma-2-9b-it, and Qwen2.5-7B-Instruct, across three settings: Direct Zero-Shot, Long-Context, and RAG. The results, summarized in Table~\ref{tab:guidelineQA_comparison}, demonstrate the definitive superiority of the RAG architecture. In nearly every instance, the RAG configuration outperforms the non-RAG baselines across the primary evaluation metrics.

These results highlight that accuracy alone is insufficient for RAG evaluation, particularly in complex information extraction tasks. On the GuidelineQA dataset, Llama-3.1-8B-Instruct achieves its highest accuracy (0.71) in the RAG setting, but its highest lexical precision (0.34) actually occurs in the Long-Context setting. This mismatch shows that standard F1-scores overly penalize minor phrasing differences, while our comprehensive list of metrics provides a more nuanced view of model behavior. In particular, our fuzzy match and List-Component F1 score capture all the essential clinical elements that a token-level F1 score may miss. By breaking down these metrics, RAG-X measures the information extraction ability of the model that the aggregate accuracy would otherwise overlook.

Finally, we observe a clear performance pattern across the generators. In the RAG setting, gemma-2-9b-it achieves the best generation quality with an F1-score of 0.56 and a semantic similarity of 0.75, reflecting its strong ability to synthesize information when guided by targeted context. In contrast, Qwen2.5-7B-Instruct struggles in the Long-Context setting, where its F1-score (0.27) and semantic similarity (0.52) dropped below its Direct Zero-shot performance. This ``lost-in-the-middle" behavior underscores the role of a retriever as it transforms the challenging information extraction from long clinical guidelines into a manageable grounded-generation task.

\noindent
\subsection{Comparison of Retrieval Models}
\noindent
To isolate the impact of the retriever, we evaluated the three retrieval models across three datasets, as summarized in Table~\ref{tab:retrieval_performance}. While Qwen3-Embedding-8B achieves the highest MRR and Recall, especially on GuidelineQA and MedQuAD-GHR, RAG-X's decoupled metrics expose important qualitative differences in information density and retrieval risk.

A key signal comes from the \textbf{Exclusive Hit Rate (EHR@1)}. On GuidelineQA, Qwen3 reaches its highest recall at $\alpha=1$, yet EHR@1 drops to 0.1, showing high information redundancy where the answer appears across multiple chunks. In contrast, MedQuAD-GHR shows higher EHR@1 values (up to 0.30), indicating a ``single-source-of-truth" pattern where the generator's success is entirely reliant on its ability to attend to the first retrieved evidence.

RAG-X also explains the ``all-or-nothing" behavior in domain-specific retrievers. On PubMedQA, MedCPT achieves the best Context Relevancy and EHR@1 despite lower recall. This confirms that, when successful, it retrieves uniquely informative evidence. Finally, the drop in Recall and EHR@1 for NV-EmbedQA-E5-v5 when moving from hybrid to pure dense search on GuidelineQA shows the importance of retaining lexical matching for clinically rigid terminology.

\begin{table*}[t]
\centering
\caption{Comprehensive retrieval performance across all three datasets and retriever configurations. Each block corresponds to a retriever model, with rows indicating the hybrid search weight ($\alpha$). We report four key retrieval metrics: MRR (Mean Reciprocal Rank), Recall@3, Ctx Rel. (LLM-judged Context Relevancy score), and EHR@1 (Exclusive hit rate for Context1).}

\label{tab:retrieval_performance}
\begin{adjustbox}{max width=\textwidth}
\begin{tabular}{lllcccccccccccc}
\toprule
\multirow{2}{*}{\textbf{Retriever}} & \multirow{2}{*}{\boldmath$\alpha$} & 
\multicolumn{4}{c}{\textbf{PubMedQA}} & 
\multicolumn{4}{c}{\textbf{GuidelineQA}} & 
\multicolumn{4}{c}{\textbf{MedQuad-GHR}} \\
\cmidrule(lr){3-6} \cmidrule(lr){7-10} \cmidrule(lr){11-14}
& & MRR & Recall & Ctx Rel. &EHR@1 & MRR & Recall & Ctx Rel. &EHR@1 & MRR & Recall & Ctx Rel. &EHR@1 \\
\midrule

BM25 & 0 & 0.31 & 0.34 & 0.44 &0.015 & 0.32 & 0.41 & 0.39 &0.14 & 0.65 & 0.77 & 0.30 &0.19 \\
\midrule

\multirow{2}{*}{Qwen3-Embedding-8B} & 0.5  &0.31  &0.34  &0.51 &0.02  &0.43  &0.54  &0.49 &0.14  &0.72  &0.81  &0.31 &0.27 \\
                          & 1.0 &0.31  &0.34  &0.52 &0.021  &0.45  &0.58  &0.52 &0.10  &0.70  &0.79  &0.30 &0.30 \\
\midrule

\multirow{2}{*}{NV-EmbedQA-E5-v5}    & 0.5  &0.31  &0.34  &0.49 &0.016  &0.39  &0.54  &0.42 &0.14  &0.66  &0.79  &0.31 &0.20 \\
                          & 1.0 &0.31  &0.34  &0.49 &0.018  &0.28  &0.41  &0.36 &0.03  &0.62  &0.74  &0.30 &0.22 \\
\midrule

\multirow{2}{*}{MedCPT-Query-Encoder}    & 0.5  &0.24  &0.31  &0.55 &0.034  &0.27  &0.34  &0.35 &0.08  &0.40  &0.54  &0.21 &0.10 \\
                          & 1.0 &0.23  &0.30  &0.50 &0.033  &0.26  &0.31  &0.32 &0.10  &0.30  &0.39  &0.18 &0.11 \\
\bottomrule
\end{tabular}
\end{adjustbox}
\label{tab:llm_retrieval_metrics}
\end{table*}

\noindent

\begin{table*}[t]
\centering
\rowcolors{2}{gray!30}{white}
\caption{Fine-grained diagnostics from our RAG-X framework with suggested actions for the best-performing Llama-3.1-8B-Instruct + Qwen3-Embedding-8B at $\alpha=1$ pipeline (MAP = 0.44) on  GuidelineQA dataset.}
\scriptsize
\begin{tabular}{l|c|p{6cm}|p{6cm}}
\hline
\textbf{Diagnostic Metric} & \textbf{Value} & \textbf{Interpretation} & \textbf{Actionable Insight} \\
\hline
Effective Use (CUE) & 49.2\% & The Only truly grounded success & System is reliable for nearly half of the queries. \\
Lucky Guess (CUE) & 33.9\% & High risk; Generator produces answers without evidence support. & Improve retriever coverage. \\
Information Blindness (CUE) & 8.5\% & Retriever was successful, but the generator missed the answer & Refine generation prompt for extraction. \\
Pairwise Redundancy (C1 and C2) & 22.0\% & Significant overlap between top ranks. & Implement Maximum Marginal Relevance (MMR) for diversity. \\
Exclusive Hit Rate (Context-2) & 6.8\% & Rank 2 rarely provides unique information. & Implement diverse re-ranking. \\
Context Adherence (LLM Judgment) & 0.84 & Generator is highly faithful to context. & No Action\\
\hline
\end{tabular}
\label{tab:llama_qwen_diagnostics}
\end{table*}

\subsection{Case Study: Diagnosing RAG Pipeline with RAG-X}
\noindent
In Table~\ref{tab:llama_qwen_diagnostics}, we demonstrate RAG-X's value using our best-performing setup (Llama-3.1-8B-Instruct + Qwen3-Embedding-8B at $\alpha=1$), which achieved the top accuracy (0.71) and MAP (0.44) on GuidelineQA. Although these metrics indicate strong performance, RAG-X's fine-grained diagnostics show that this success is partly misleading.

\noindent
\textbf{The Accuracy Fallacy (CUE Analysis):} RAG-X shows that much of the system's apparent success is not truly grounded. Although 49.2\% of responses reflect \textit{Effective Use}, a substantial 33.9\% are \textit{Lucky Guess}, where the retriever misses the answer but the model still generates a confident, correct-looking response. Without RAG-X, this behavior would be hidden under the overall Accuracy score.

\noindent
\textbf{Quantifying Retrieval Waste:} Although standard Recall (57.6\%) suggests adequate coverage, RAG-X uncovers a major inefficiency, with 22.0\% Pairwise redundancy between the top contexts and only 6.8\% Exclusive Hit Rate at Rank 2. Practically, the retriever is returning overlapping evidence rather than complementary context, wasting retrieval capacity.

\noindent
\textbf{Diagnostic of Attribution Error:} RAG-X reveals a 14\% gap between Accuracy (71\%) and overall Context Hit Rate (57.6\%), showing that a few correct answers come from the model's parametric knowledge rather than retrieved evidence. The high Context Adherence Score (0.84) then incorrectly attributes these answers to the context, demonstrating that adherence alone can give a false sense of grounding highlighting the critical role of CUE metric in making these errors visible. 

\noindent

\noindent

\section{Conclusion}
\noindent
The safe integration of AI in high-stakes medical domains depend on RAG systems that are not only capable but verifiably grounded. This work introduced RAG-X, a diagnostic framework designed to move beyond aggregate accuracy by offering a component-level X-ray of the RAG pipeline. By breaking down the system behavior through the CUE framework, we show that high accuracy can be misleading, often masking grounding failures and deceptive lucky guesses that standard benchmarks overlook. As RAG systems become increasingly integrated into clinical workflows, the diagnostic transparency offered by RAG-X becomes critical for ensuring safety and trustworthiness. This work provides the targeted validation needed to transform LLMs from general-purpose assistants to reliable, evidence-based clinical tools, thereby setting a stronger standard for responsible medical AI systems.

\bibliographystyle{IEEEtran}
\bibliography{references.bib}

@article{gaber2025,
  author = {Gaber, Farieda and Shaik, Maqsood and Allega, Fabio and Bilecz, Agnes Julia and Busch, Felix and Goon, Kelsey and others},
  title = {Evaluating Large Language Model Workflows in Clinical Decision Support for Triage and Referral and Diagnosis},
  volume = {8},
  pages = {263},
  year = {2025},
  doi = {10.1038/s41746-025-01684-1},
  journal = {npj Digital Medicine},
}

@inproceedings{wang-etal-2025-towards-adapting,
    title = "Towards Adapting Open-Source Large Language Models for Expert-Level Clinical Note Generation",
    author = "Wang, Hanyin  and
      Gao, Chufan  and
      Liu, Bolun  and
      Xu, Qiping  and
      Hussein, Guleid  and
      Labban, Mohamad El  and
      others",
    booktitle = "Findings of the Association for Computational Linguistics: ACL 2025",
    year = "2025",
    pages = "12084--12117",
}

@article{llm-med-edu2023,
  author = {Abd-alrazaq, Alaa and AlSaad, Rawan and Alhuwail, Dari and Ahmed, Arfan and Healy, Padraig Mark and Latifi, Syed and others},
  title = {Large Language Models in Medical Education: Opportunities, Challenges, and Future Directions},
  journal = {JMIR Medical Education},
  volume = {9},
  pages = {e48291},
  year = {2023},
  doi = {10.2196/48291},
}

@article{survey_hallucination,
author = {Ji, Ziwei and Lee, Nayeon and Frieske, Rita and Yu, Tiezheng and Su, Dan and Xu, Yan and others},
title = {Survey of Hallucination in Natural Language Generation},
year = {2023},
volume = {55},
number = {12},
doi = {10.1145/3571730},
journal = {ACM Comput. Surv.},
articleno = {248},
}

@inproceedings{rag_4_nlp,
author = {Lewis, Patrick and Perez, Ethan and Piktus, Aleksandra and Petroni, Fabio and Karpukhin, Vladimir and Goyal, Naman and others},
title = {Retrieval-augmented generation for knowledge-intensive {NLP} tasks},
booktitle = {Proc. NeurIPS},
articleno = {793},
year = {2020},
}

@inproceedings{xiong-etal-2024-benchmarking,
    title = "Benchmarking Retrieval-Augmented Generation for Medicine",
    author = "Xiong, Guangzhi  and
      Jin, Qiao  and
      Lu, Zhiyong  and
      Zhang, Aidong",
    booktitle = "Findings of the Association for Computational Linguistics: ACL 2024",
    year = "2024",
    doi = "10.18653/v1/2024.findings-acl.372",
    pages = "6233--6251",
}

@article{zheng2023judging,
  title={Judging LLM-as-a-Judge with MT-Bench and Chatbot Arena},
  author={Zheng, Lianmin and Chiang, Wei-Lin and Sheng, Ying and Zhuang, Siyuan and Wu, Zhanghao and Zhuang, Yonghao and others},
  journal={Advances in neural information processing systems},
  volume={36},
  pages={46595--46623},
  year={2023}
}

@article{luo2019hybrid,
  title={A hybrid normalization method for medical concepts in clinical narrative using semantic matching},
  author={Luo, Yen-Fu and Sun, Weiyi and Rumshisky, Anna},
  journal={AMIA Summits on Translational Science Proceedings},
  pages={732},
  year={2019}
}

@misc{li2024scoping,
  title={A scoping review of using large language models ({LLMs}) to investigate electronic health records ({EHRs})},
  author={Li, Lingyao and Zhou, Jiayan and Gao, Zhenxiang and Hua, Wenyue and Fan, Lizhou and Yu, Huizi and others},
  year={2024},
  note= {arXiv:2405.03066 [cs.CL]},
  url= {https://arxiv.org/abs/2405.03066}
}

@misc{dubey2024llama,
  title={The {Llama} 3 herd of models},
  author={Dubey, Abhimanyu and Jauhri, Abhinav and Pandey, Abhinav and Kadian, Abhishek and Al-Dahle, Ahmad and Letman, Aiesha and others},
  note={arXiv:2407.21783 [cs.AI]},
  doi={10.48550/arXiv.2407.21783},
  url={https://arxiv.org/abs/2407.21783},
  year={2024}
}

@misc{gemma_2024,
  author    = {{Google Research}},
  title     = {Gemma 2 9B-IT Model Card},
  year      = {2024},
  howpublished = {\url{https://huggingface.co/google/gemma-2-9b-it}},
  note      = {[Online; accessed Jan. 30, 2026]}
}

@misc{qwen2,
  author    = {Yang, An and others},
  title     = {Qwen2 Technical Report},
  year      = {2024},
  note      = {arXiv:2407.10671 [cs.CL]},
  url       = {https://arxiv.org/abs/2407.10671}
}

@misc{deepseekai2024deepseekv3technicalreport,
  author    = {{DeepSeek-AI}},
  title     = {DeepSeek-{V3} Technical Report},
  year      = {2024},
  note      = {arXiv:2412.19437 [cs.CL]},
  url       = {https://arxiv.org/abs/2412.19437}
}

@inproceedings{krishna-etal-2025-fact,
    title = "Fact, Fetch, and Reason: A Unified Evaluation of Retrieval-Augmented Generation",
    author = "Krishna, Satyapriya  and
      Krishna, Kalpesh  and
      Mohananey, Anhad  and
      Schwarcz, Steven  and
      Stambler, Adam  and
      Upadhyay, Shyam  and
      Faruqui, Manaal",
    booktitle = "Proc. NAACL",
    year = "2025",
    doi = "10.18653/v1/2025.naacl-long.243",
    pages = "4745--4759",   
}

@inproceedings{saadfalcon2024ares,
  author={J. Saad-Falcon and O. Khattab and C. Potts and M. Zaharia},
  title={{ARES: An Automated Evaluation Framework for Retrieval-Augmented Generation Systems}},
  booktitle={Proc. NAACL},
  year={2024},
  doi={10.48550/arXiv.2311.09476},
}

@article{luo2024zero,
  title={Zero-shot learning to extract assessment criteria and medical services from the preventive healthcare guidelines using large language models},
  author={Luo, Xiao and Tahabi, Fattah Muhammad and Marc, Tressica and Haunert, Laura Ann and Storey, Susan},
  journal={Journal of the American Medical Informatics Association},
  volume={31},
  number={8},
  pages={1743--1753},
  year={2024},
}

@article{levra2025large,
  title={A large language model-based clinical decision support system for syncope recognition in the emergency department: A framework for clinical workflow integration},
  author={Levra, Alessandro Giaj and Gatti, Mauro and Mene, Roberto and Shiffer, Dana and Costantino, Giorgio and Solbiati, Monica and others},
  journal={European journal of internal medicine},
  volume={131},
  pages={113--120},
  year={2025},
}

@misc{ngo2024medrgb,
  author = {Ngo, Nghia Trung and Nguyen, Chien Van and Dernoncourt, Franck and Nguyen, Thien Huu},
  title = {Comprehensive and Practical Evaluation of Retrieval-Augmented Generation Systems for Medical Question Answering},
  year = {2024},
  note      = {arXiv:2411.09213 [cs.CL]},
  doi       = {10.48550/arXiv.2411.09213},
  url       = {https://arxiv.org/abs/2411.09213}
}

@inproceedings{lu2024clinicalrag,
  title={{ClinicalRAG}: enhancing clinical decision support through heterogeneous knowledge retrieval},
  author={Lu, Yuxing and Zhao, Xukai and Wang, Jinzhuo},
  booktitle={Proc. KnowLLM 2024},
  pages={64--68},
  year={2024}
}

@article{improving_LLM_meta_analysis,
    author = {Liu, Siru and McCoy, Allison B and Wright, Adam},
    title = {Improving large language model applications in biomedicine with retrieval-augmented generation: a systematic review, meta-analysis, and clinical development guidelines},
    journal = {Journal of the American Medical Informatics Association},
    volume = {32},
    number = {4},
    pages = {605-615},
    year = {2025},
    doi = {10.1093/jamia/ocaf008},
}

@article{enhancing_med_AI,
  author = {Gargari, Omid Kohandel and Habibi, Gholamreza},
  title = {Enhancing Medical {AI} with Retrieval-Augmented Generation: A Mini Narrative Review},
  journal = {Digital Health},
  volume = {11},
  pages = {20552076251337177},
  year = {2025},
  doi = {10.1177/20552076251337177},
}

@inproceedings{es-etal-2024-ragas,
    title = "{RAGA}s: Automated Evaluation of Retrieval Augmented Generation",
    author = "Es, Shahul  and
      James, Jithin  and
      Espinosa Anke, Luis  and
      Schockaert, Steven",
    booktitle = "Proc. EACL",
    year = "2024",
    doi = "10.18653/v1/2024.eacl-demo.16",
    pages = "150--158",
}

@misc{kim2025medical,
  title={Medical hallucinations in foundation models and their impact on healthcare},
  author={Kim, Yubin and others},
  note= {arXiv:2503.05777 [cs.CL]},
  url= {https://arxiv.org/abs/2503.05777},
  year={2025}
}

@article{clusmann2023future,
  title={The future landscape of large language models in medicine},
  author={Clusmann, Jan and Kolbinger, Fiona R and Muti, Hannah Sophie and Carrero, Zunamys I and Eckardt, Jan-Niklas and Laleh, Narmin Ghaffari and others},
  journal={Communications medicine},
  volume={3},
  number={1},
  pages={141},
  year={2023},
}

@misc{wu2024well,
  title={How well do {LLMs} cite relevant medical references? {An} evaluation framework and analyses},
  author={Wu, Kevin and others},
  note= {arXiv:2402.02008 [cs.CL]},
  url= {https://arxiv.org/abs/2402.02008},
  year={2024}
}

@article{benabacha2019question,
  author    = {Ben Abacha, Asma and Demner-Fushman, Dina},
  title     = {A Question-Entailment Approach to Question Answering},
  journal   = {BMC Bioinformatics},
  volume    = {20},
  number    = {1},
  pages     = {511},
  year      = {2019},
  doi       = {10.1186/s12859-019-3119-4},
}

@inproceedings{jin2019pubmedqa,
  author       = {Jin, Qiao and Dhingra, Bhuwan and Liu, Zhengping and Cohen, William W. and Lu, Xinghua},
  title        = {{PubMedQA}: A Dataset for Biomedical Research Question Answering},
  booktitle = {Proc. EMNLP-IJCNLP},
  pages     = {2567--2577},
  year      = {2019},
  doi       = {10.18653/v1/D19-1259}
}

@misc{qwen3embedding,
  title={Qwen3 Embedding: Advancing Text Embedding and Reranking Through Foundation Models},
  author={Zhang, Yanzhao and others},
  note= {arXiv:2506.05176 [cs.CL]},
  url= {https://arxiv.org/abs/2506.05176},
  year={2025}
}

@article{jin2023medcpt,
  title={{MedCPT}: Contrastive Pre-trained Transformers with large-scale PubMed search logs for zero-shot biomedical information retrieval},
  author={Jin, Qiao and others},
  journal={Bioinformatics},
  volume={39},
  number={11},
  pages={btad651},
  year={2023},
}

@misc{weaviate,
  author       = {Weaviate},
  title        = {Weaviate: An open-source vector search engine},
  year         = {2024},
  howpublished = {\url{https://weaviate.io}},
  note         = {Accessed: 2026-01-30}
}

@misc{cohen2025ragxplain,
  author       = {Cohen, Dvir and Burg, Lin and Barkan, Gilad},
  title        = {{RAGXplain}: From Explainable Evaluation to Actionable Guidance of RAG Pipelines},
  year         = {2025},
 note      = {arXiv:2505.13538 [cs.IR]},
  doi       = {10.48550/arXiv.2505.13538},
  url       = {https://arxiv.org/abs/2505.13538}
}

@misc{nvidia2024nvembedqa,
  author       = {NVIDIA},
  title        = {NVIDIA NIM: nv-embedqa-e5-v5},
  year         = {2024},
  howpublished = {\url{https://docs.api.nvidia.com/nim/reference/nvidia-nv-embedqa-e5-v5}},
  note         = {Accessed: 2026-01-31}
}

@misc{sentence-transformers-pretrained-models,
  author       = {Sentence Transformers},
  title        = {Pretrained Models — Sentence Transformers Documentation},
  year         = {2024},
  howpublished = {\url{https://www.sbert.net/docs/sentence_transformer/pretrained_models.html}},
  note         = {Accessed: 2026-01-31}
}

\end{document}